\documentclass[conference]{IEEEtran}

\usepackage{booktabs}
\usepackage{amsmath,amssymb,amsfonts}
\usepackage{algorithmic}
\usepackage{graphicx}
\usepackage{textcomp}
\usepackage{xcolor}
\usepackage{biblatex}
\def\BibTeX{{\rm B\kern-.05em{\sc i\kern-.025em b}\kern-.08em
    T\kern-.1667em\lower.7ex\hbox{E}\kern-.125emX}}
\begin{document}

\title{PICOs-RAG: PICO-supported Query Rewriting for Retrieval-Augmented Generation in Evidence-Based Medicine
}

\author{\IEEEauthorblockN{1\textsuperscript{st}Mengzhou Sun}
\IEEEauthorblockA{\textit{Faculty of Computing} \\
\textit{Harbin Institute of Technology}\\
Harbin, China \\
mzsun@ir.hit.edu.cn}
\and
\IEEEauthorblockN{2\textsuperscript{nd} Sendong Zhao}
\IEEEauthorblockA{\textit{Faculty of Computing} \\
\textit{Harbin Institute of Technology}\\
Harbin, China\\
sdzhao@ir.hit.edu.cn
}
\and
\IEEEauthorblockN{3\textsuperscript{rd}Jianyu Chen}
\IEEEauthorblockA{\textit{Faculty of Computing} \\
\textit{Harbin Institute of Technology}\\
Harbin, China \\
hcwang@ir.hit.edu.cn
    }
\and
\IEEEauthorblockN{4\textsuperscript{th}Bing Qin}
\IEEEauthorblockA{\textit{Faculty of Computing} \\
\textit{Harbin Institute of Technology}\\
Harbin, China \\
qinb@ir.hit.edu.cn
    }
}

\maketitle

\begin{abstract}

Evidence-based medicine (EBM) research has always been of paramount importance. It is important to find appropriate medical theoretical support for the needs from physicians or patients to reduce the occurrence of medical accidents. This process is often carried out by human querying relevant literature databases, which lacks objectivity and efficiency. Therefore, researchers utilize retrieval-augmented generation (RAG) to search for evidence and generate responses automatically. However, current RAG methods struggle to handle complex queries in real-world clinical scenarios. For example, when queries lack certain information or use imprecise language, the model may retrieve irrelevant evidence and generate unhelpful answers. To address this issue, we present the PICOs-RAG to expand the user queries into a better format. Our method can expand and normalize the queries into professional ones and use the PICO format, a search strategy tool present in EBM, to extract the most important information used for retrieval. This approach significantly enhances retrieval efficiency and relevance, resulting in up to an 8.8\% improvement compared to the baseline evaluated by our method. Thereby the PICOs-RAG improves the performance of the large language models into a helpful and reliable medical assistant in EBM.

\end{abstract}


\section{Introduction}


As evidence-based medicine (EBM) becomes increasingly widespread, physicians recognize that clinical diagnoses supported by substantial medical evidence tend to have lower misdiagnosis rates. With the growing amount of medical literature, each diagnosis now requires extensive supporting evidence. Consequently, medical professionals are turning to AI as an auxiliary tool to implement EBM effectively. However, AI must be capable of integrating all patient information and the available supporting evidence \cite{clusmann2023future}. Traditional models, due to memory limitations and other constraints, have struggled to manage and synthesize this large amount of evidence comprehensively ~\cite{friedman2013natural,nadkarni2011natural}. This situation persists until the advent of large language models (LLMs). With the enhancement of various medical LLMs and continuous optimization of training methods, many of these models have now reached the capability to effectively perform this critical clinical task \cite{friedman1999natural}.

\begin{figure}
    \centering
    \includegraphics[width=1.0\linewidth]{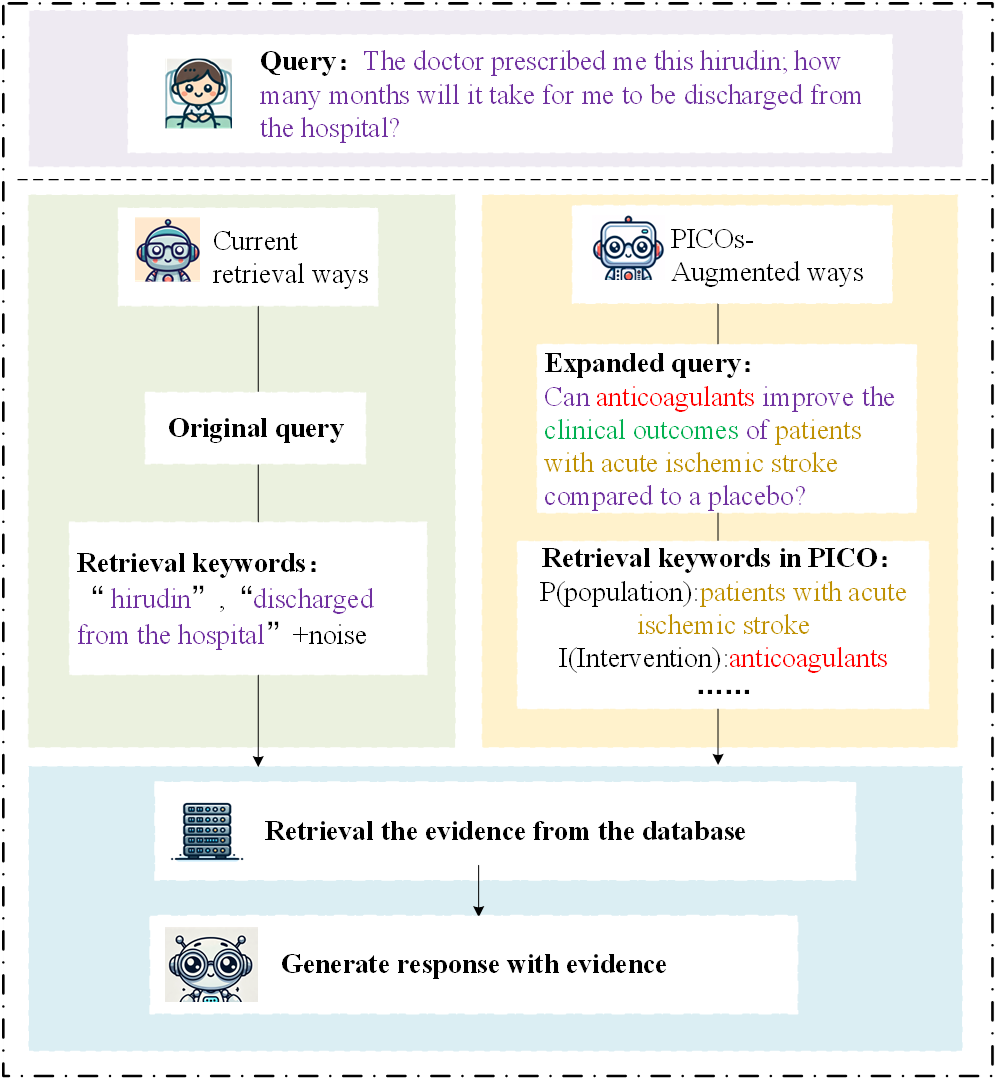}
    \caption{The architecture of current RAG methods compared with our PICOs method. The PICOs enhance the relevance of the evidence retrieval process and improve the accuracy of responses, enabling the LLM to provide precise and targeted assistance.}
    \label{fig:1}
\end{figure}


Recently, LLMs have undergone multiple rounds of reinforcement and improvement~\cite{alam2023automated}. They can utilize the retrieval-augmented generation (RAG) method to significantly reduce the hallucination to the levels accepted by many professional researchers ~\cite{zhang2023siren,huanglei2023survey}. However, the LLMs still have many limitations for their application in medicine. Current LLMs still can not answer clearly in the domain such as clinical medicine for the complicated queries. From real medical question-and-answer Platforms~\cite{he2019applying}, we find that due to the lack of professional knowledge, users such as patients may use non-professional words and omit crucial information when they query doctors. In this situation, RAG frequently fails to retrieve the most critical evidence and gets a noisy index. As shown in Figure \ref{fig:1}, the current RAG retrieves the keywords like `hirudin', `discharged from hospital', and other noise, which may be hard to retrieve the critical evidence. Therefore, the LLMs may generate an irrelevant answer which is unhelpful for users. This result significantly diminishes the user experience, which decreases trust in LLMs.



To address the above problem, we propose a method, PICOs-RAG, based on LLMs for query expanding and extracting the critical information from the query. As shown in the right part of Figure \ref{fig:1}, our approach is divided into three steps: expanding the query, extracting the PICOs, and retrieving the evidence. We first expand and normalize the query and then extract by a PICO format. The words are separated into four main components: Population, Intervention, Comparison, and Outcome~\cite{methley2014pico}. These components contain the core information of the query. By providing the retrieved evidence along with the expanded query to the LLM, we obtain more accurate and professional responses compared to current RAG methods.

We also employ two different evaluation methods to assess the accuracy and relevance of the model-generated responses. Our findings indicate that PICOs-RAG provides a stable and significant improvement across the entire EBM RAG system. Specifically, the accuracy increased by 6.2\% under the accuracy evaluation method, and by 8.6\% under the relevance evaluation method.

Our approach primarily optimizes the RAG model for EBM tasks by addressing the issue of weak relevance in queries and evidence. The main contributions of our method are as follows:    
\begin{itemize}
    \item  We enhance the feasibility and effectiveness of user input by expanding and normalizing queries based on medical specialties.
    \item     We present a PICO-based method to extract the core elements of the question, aiding the subsequent RAG process in more targeted evidence retrieval and response generation.

    \item    We propose two different evaluation approaches to evaluate medical responses, assessing LLM outputs more comprehensively and robustly.
\end{itemize}

\section{Background and related works}
\subsection{Evidence Based Medicine and PICO}

EBM, as an essential practice method in modern medicine, aims to make optimal clinical decisions by integrating the best research evidence, clinical expertise, and patient preferences \cite{subbiah2023next,armeni2022digital}. In recent years, the development of natural language processing and the advancements in data processing have greatly promoted the growth and application of EBM \cite{vaid2024generative}. However, EBM still faces many challenges and limitations in practical application. The core principle of EBM is to emphasize high-quality research evidence to guide clinical practice. This is crucial for improving the scientific and effective nature of medical decision-making. EBM combines systematic research evidence, clinical expertise, and patient values to provide the most appropriate clinical care. Through systematic reviews and meta-analyses, physicians can access the latest and most reliable medical research results, enabling them to make more scientifically informed treatment decisions.~\cite{borenstein2021introduction}
\begin{figure}
    \centering
    \includegraphics[width=1.0\linewidth]{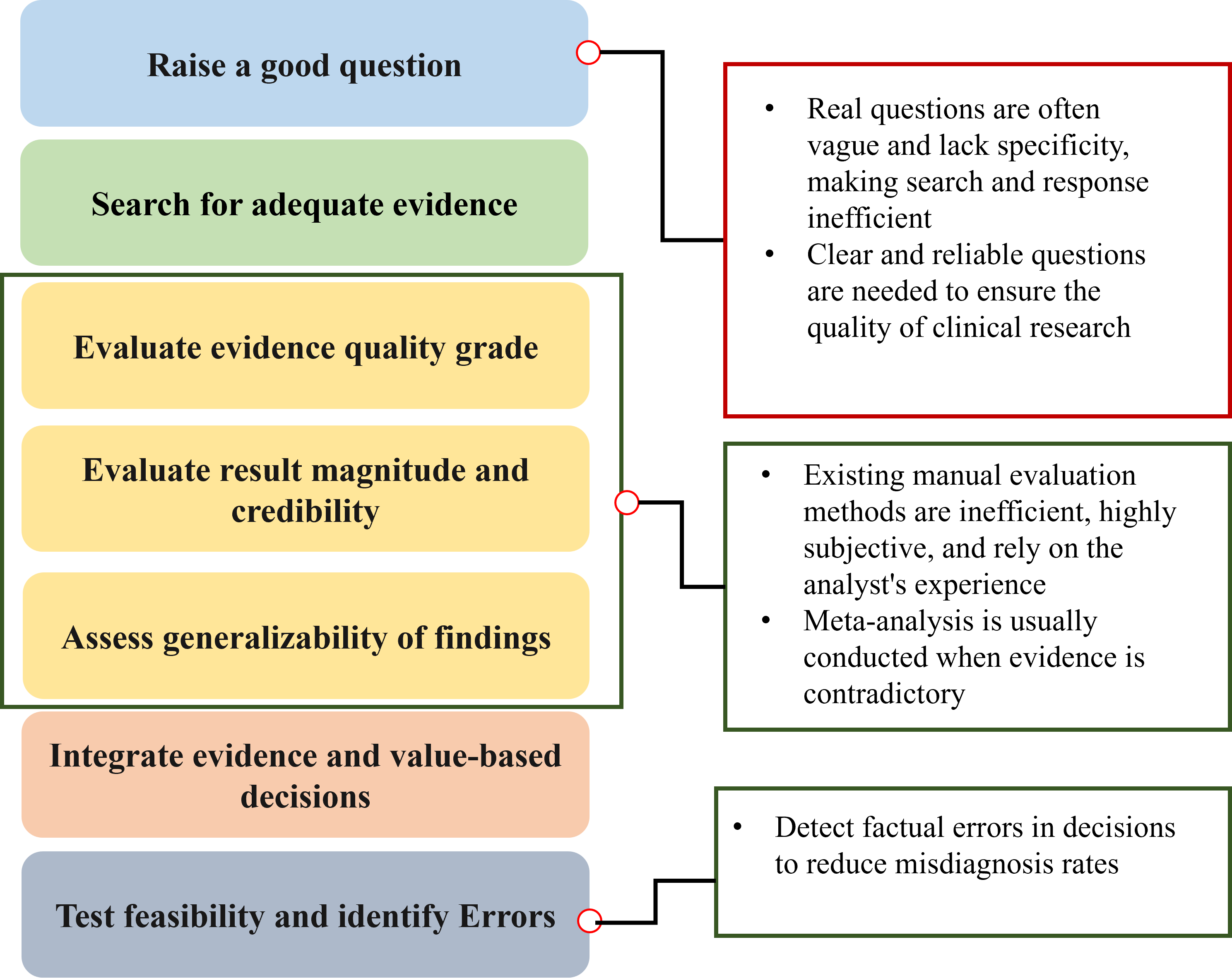}
    \caption{Steps of EBM. EBM primarily involves formulating the queries, searching for all possible relevant evidence, filtering and reordering the obtained evidence, and answering the question based on the evidence gathered. Our method focuses on optimizing the query, addressing the issue where queries may lack professionalism and clarity due to insufficient knowledge~\cite{methley2014pico}.}
    \label{fig:2}
\end{figure}

As shown in Figure \ref{fig:2}, EBM can be divided into six parts. Currently, there is a high demand for AI applications in EBM. As stated in the work of Subbish~\cite{subbiah2023next}, AI should comprehensively utilize all patient information, including extracting, organizing, and mining vast amounts of historical data, genomics, and other omics analyses, as well as all published clinical studies, to provide the next generation of evidence for deep medicine. traditional models like BERT and Transformer, due to memory limitations and the capabilities of physicians, cannot fully integrate these contents, necessitating larger models for comprehensive integration and decision-making.

Research on the effectiveness of medical training~\cite{ngusie2023effect} has indicated that current EBM training can significantly improve EBM knowledge and skills of healthcare professionals and promote their application in clinical practice. This is of great significance for improving the quality of healthcare services. However, there are differences in the acceptance and adaptability of training among different healthcare professionals, which may lead to variability in training outcomes among individuals. Further research is needed to adjust training content and methods based on individual differences to enhance overall effectiveness.
PICO is a concept in EBM used to identify the components of clinical evidence within a systematic review.~\cite{hu2023towards} This concept effectively describes the primary components of a medical text. We apply this concept and methodology to the RAG-LLMs approach to extract the most critical parts of medical texts, and we termed this enhanced method PICOs. The PICO tool focuses on the four main concepts of ``Population, Intervention, Comparison, and Outcome" within Medical articles. Moreover, the structured approach of PICO facilitates the efficient screening and evaluation of a vast amount of literature, ensuring the relevance and quality of the selected studies.
\subsection{RAG for EBM}

LLMs have made significant strides in natural language processing in recent years. Models such as GPT-4o and Claude have demonstrated exceptional analytical and processing capabilities, leading to substantial breakthroughs in fields like medicine, military, and law. Currently, LLMs can adapt to the latest medical research and clinical practices, enhancing decision-making quality and accuracy. As shown in the Google work~\cite{singhal2023large} LLMs can be applied to various medical domains. In terms of methodology, researchers have developed the RAG approach, which combines retrieval and generation to reduce hallucinations. In rapidly evolving fields such as medicine, researchers often express concerns that LLMs might provide outdated responses due to their static internal knowledge base. RAG effectively mitigates this issue by integrating up-to-date information. The concept of EBM, which relies on extensive medical evidence as the basis for judgment, aligns perfectly with the RAG approach. Thus, the RAG method is highly suitable for current EBM practices. It also represents an effective application of LLMs to assist clinicians in solving clinical problems.

However, the application of LLMs in the medical field presents its challenges. Firstly, at the application level, the questions addressed in EBM are often posed by patients colloquially or informally. Directly using these questions to search for evidence often results in irrelevant or even meaningless search results. After retrieving relevant evidence, researchers in EBM frequently find an overwhelming amount of evidence, some of which may be contradictory, necessitating meta-analysis. This meta-analysis process is time-consuming, labor-intensive, and highly subjective. Analyzing these pieces of evidence using LLMs can mislead the models into making incorrect judgments. The third issue is the inherent limitations of LLMs; the output often contains factual errors. Although incorporating the RAG process has improved this, models sometimes rely on their internal knowledge and overlook more reliable information from the RAG process. Therefore, a means of verifying factual errors is necessary. Addressing these issues, this paper focuses on optimizing the first problem by starting with user queries. Using the PICO principles of EBM, we formalize the original input, making the retrieval process more efficient and the questions more targeted.

\subsection{Query expanding to augmenting RAG}

When applying NLP models to the data in the professional domain, many researchers have noted that not all user inputs are complete and clear. For example, in online medical Q\&A data, users often lack relevant medical knowledge, leading to unfocused and unprofessional questions. This can cause misunderstandings of medical concepts or overly verbose questions. These questions will fail to efficiently convey the user's current intent. This issue significantly impacts both the retrieval process and the answer generation process in RAG models, greatly limiting the stable performance on such datasets.

To address this challenge, numerous efforts have been made to rewrite the input queries of LLMs, making them more targeted and professional, thereby improving the accuracy of the responses. Current research approaches mainly fall into two categories: using LLMs and small models to expand the input. In the method of query rewriting~\cite{ma2023query}, the authors compared three different query schemes, fully demonstrating the effectiveness of current query rewriting methods. In the work RAFE~\cite{mao2024rafe}, users effectively validated the efficacy of query rewriting through comparative experiments on different datasets and SFT methods.

These two works have inspired this paper. However, in our experiments, we find that directly applying these methods fails to effectively handle query expansion in the medical field. Our analysis reveals two main reasons for this failure. Firstly, the medical domain datasets are too specialized, presenting a knowledge gap that small models cannot bridge. Simple-term translation offers little help to RAG. Secondly, relying solely on LLMs for expansion results in excessively long retrieval content, leading to retrieval results filled with irrelevant information.

To overcome these shortcomings, we propose a novel method that combines large and small models for query rewriting based on the PICOs criteria. This approach aims to enhance the precision and relevance of queries, thereby improving the performance of RAG models in the medical domain.
\section{Methods}

\subsection{Task definition}\label{3.1}
\begin{figure}
    \centering
    \includegraphics[width=1.0\linewidth]{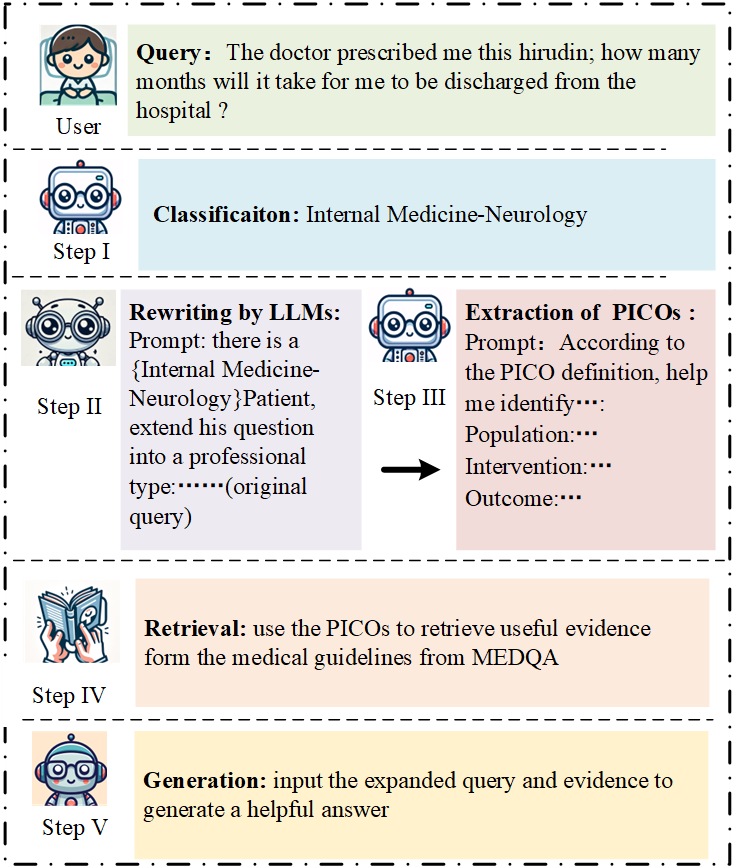}
    \caption{PICOs method to help RAG used on EBM. PICOs first capture the objective aspects of the queries. The classification results then prompt the LLM to expand and refine the query. After the query is completed, we use the LLM to extract the PICO elements from the question. These PICO elements are then passed to the retrieval module for more precise extraction. The retrieved results, along with the refined question, are then fed into the response model to generate the final response.}
    \label{fig:3}
\end{figure}



Our task is to take the user's medical queries as input and provide answers. To align with the principles of EBM, we aim to offer not only a convincing response but also the corresponding evidence to support the given answer.

In this task, our method involves several steps as shown in Figure \ref{fig:3}. First, we classify the query into different medical disciplines. Specifically, we get the classification results and then expand the query based on them. Afterward, we extract the core PICOs components from the expanded content. We then use these extracted PICOs to perform targeted evidence retrieval from a knowledge base composed of medical guidelines and literature. Finally, we submit the retrieved evidence along with the user's expanded and normalized query to the target LLM, which generates the final response.

\subsection{Query classification}\label{4.2}

We first classify the queries by the medical disciplines. Expanding queries requires a certain medical prompt to prevent the model from generating Rambling content, which is quite harmful to the next PICO extraction. In clinical practice, different disciplines focus on different parts of patients' information. For instance, in internal medicine, doctors tend to ask detailed questions about dietary habits, medical history, and family history. These symptoms are always important clues for patients if they have problems with internal medicine. Therefore, identifying the relevant disciplines can significantly aid the model in effectively expanding the user's query. We have made an experiment and proved this result in section IV.C.

\begin{figure}
    \centering
    \includegraphics[width=0.8\linewidth]{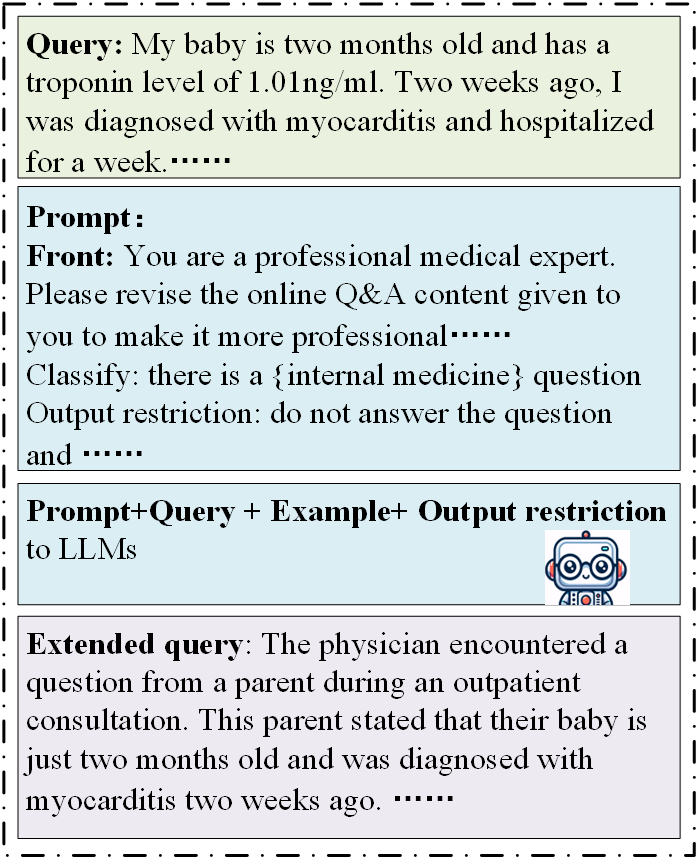}
    \caption{Examples of the method we expand and normalize query. The sentences rewritten by the LLM use more professional language, and the user's query is effectively completed.}
    \label{fig:4}
\end{figure}
\subsection{Query expanding}\label{4.2}
After the classification, we utilize the LLMs for normalizing and expanding the queries. As shown in Figure \ref{fig:4}. The LLMs normalize and expand the query following the categories of the disciplines. We provide the model with the required response format and prompts about the relevant medical discipline. This information helps the model understand the necessary content for expansion, making its output more accurate and targeted.

In this step, since we choose not to impose length and output restrictions on the expanding content, we observed that the model occasionally directly answers the query after completing the expansion. This phenomenon can interfere with subsequent PICOs extraction. To address this, we incorporate output rules into the prompt to guide the model's output, making it easier to filter out any irrelevant or unauthorized content from the model's response.


\subsection{PICOs extraction}\label{4.2}
\begin{figure}
    \centering
    \includegraphics[width=0.8\linewidth]{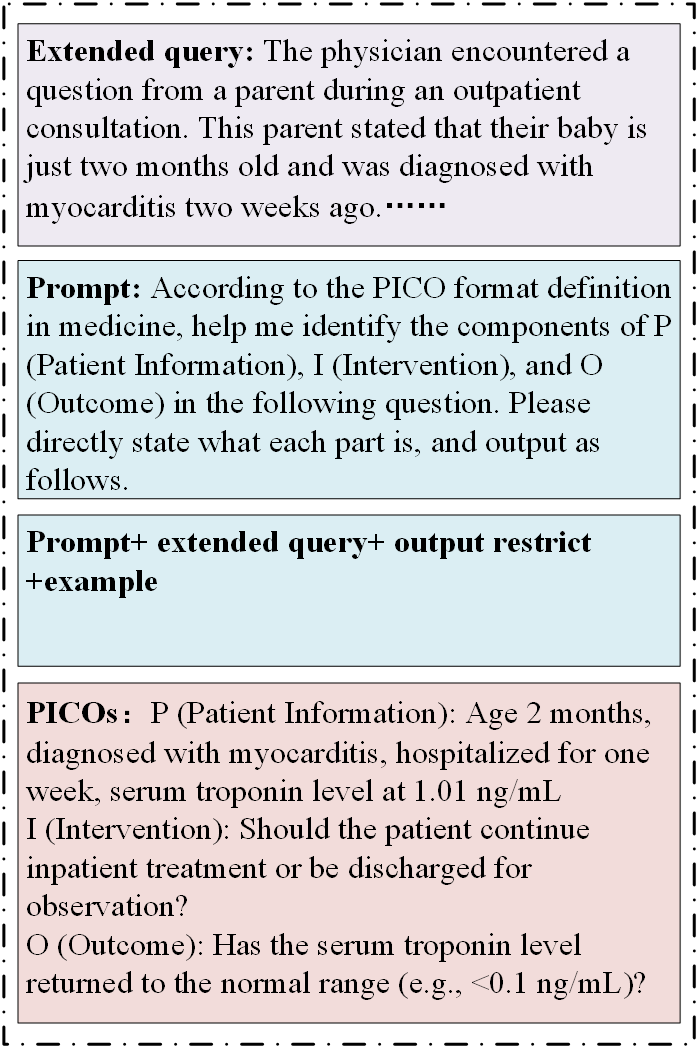}
    \caption{Examples of the method we extract the PICOs from the rewritten query. The PICO-formatted text extracted through PICOs is more targeted compared to the expanded sentences, resulting in content retrieval that is more closely aligned with the query.}
    \label{fig:5}
\end{figure}

Upon using the LLMs, we find that the outputs still contain a substantial amount of irrelevant information. Directly using these irrelevant contents for extraction significantly reduces our retrieval efficiency. Therefore, as shown in Figure \ref{fig:5}, we adapt the PICOs method to filter the expanding LLM output using the PICO format. By extracting the PICO (Population, Intervention, Comparison, Outcome) information, we eliminate other unnecessary information from the content. 

After we obtain the required PICO information, we use these core words for evidence retrieval. These PICO elements are vectorized by LLMs and then compared with the vectors in the index, which is present in MED-PaLM2 work \cite{singhal2023towards} and filled with medical guidelines. The content with the highest relevance is selected for retrieval. The filtered content is more targeted and efficient, and this method ultimately proved to enhance the RAG retrieval capability significantly. Once we have obtained high-quality evidence, we provide this evidence along with the expanded query to the response model, which then generates the final answer.

\subsection{Answer evaluation}\label{4.2}
Finally, we need to evaluate the quality of our model's outputs. In open-ended medical Q\&A tasks, this evaluation is typically challenging. Direct evaluation by LLMs tends to be highly unstable, and manual assessment is time-consuming and labor-intensive. Fortunately, we can access the best answers selected by the users who posed the questions, which serve as a valuable reference. To assess the accuracy and relevance of the model's responses, we employ two different evaluation methods: 1)Method A: We use an LLM to compare the model's responses with the best answers to ensure that the response includes the semantic information contained in the best answer. this method is referred to as Method A. 2)Method B: We use a more comprehensive and knowledgeable LLM to evaluate the overall accuracy of the model's response, which is considered as Method B. These two evaluation metrics allow for a more thorough interpretation of the quality of generated model responses for the users.

\begin{figure}
    \centering
    \includegraphics[width=1.0\linewidth]{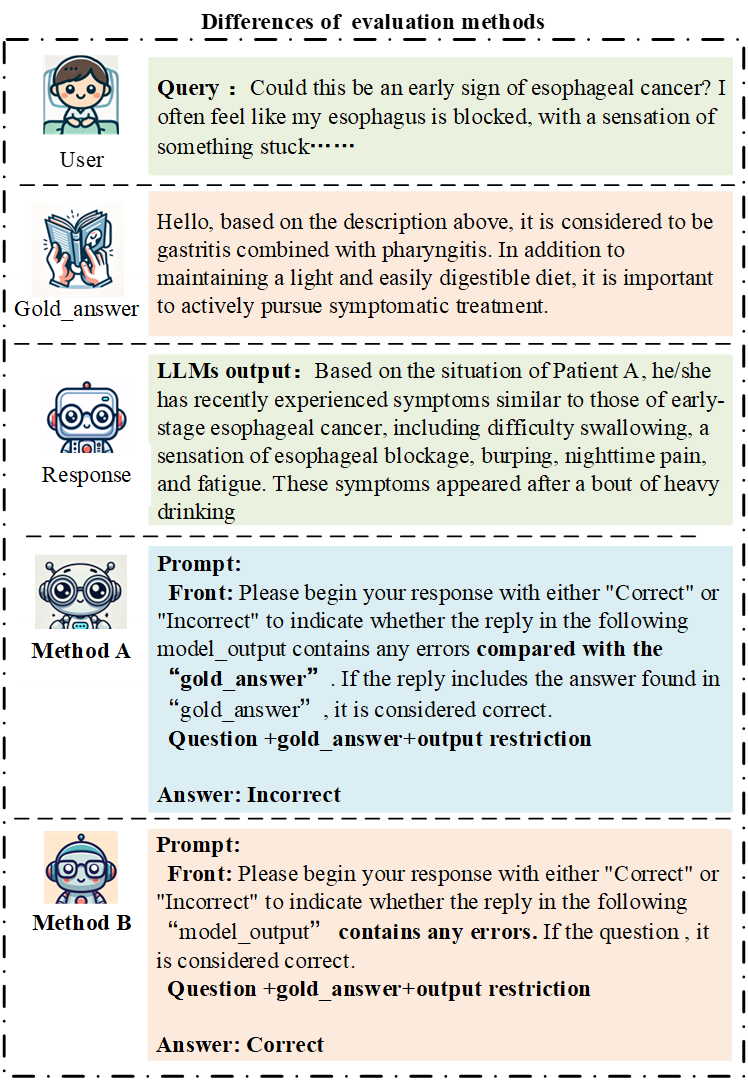}
    \caption{Examples of the two methods of evaluation. We utilize different prompts, Method A and Method B, to evaluate the generated responses regarding relevance to the gold standard and total knowledge accuracy. This approach allows for a more comprehensive assessment of the responses within medical contexts.}
    \label{fig:6}
\end{figure}

\section{Experiments and results}

\subsection{Experiments setup}\label{4.2}

Firstly, we explore various medical question datasets to simulate real-world queries from non-professional users. We find there are adequate datasets in English while current LLMs are highly trained on these datasets. Therefore, Even without any additional support, these LLMs may achieve high accuracy in English datasets. To better demonstrate the effectiveness of our method, we ultimately choose a Chinese dataset named 
webMedQA~\cite{he2019applying}. It comprises Q\&A pairs from online medical forums. This dataset features many conversational and non-specialist questions in Chinese, making it suitable for our study. 

\begin{figure}
    \centering
    \includegraphics[width=1.0\linewidth]{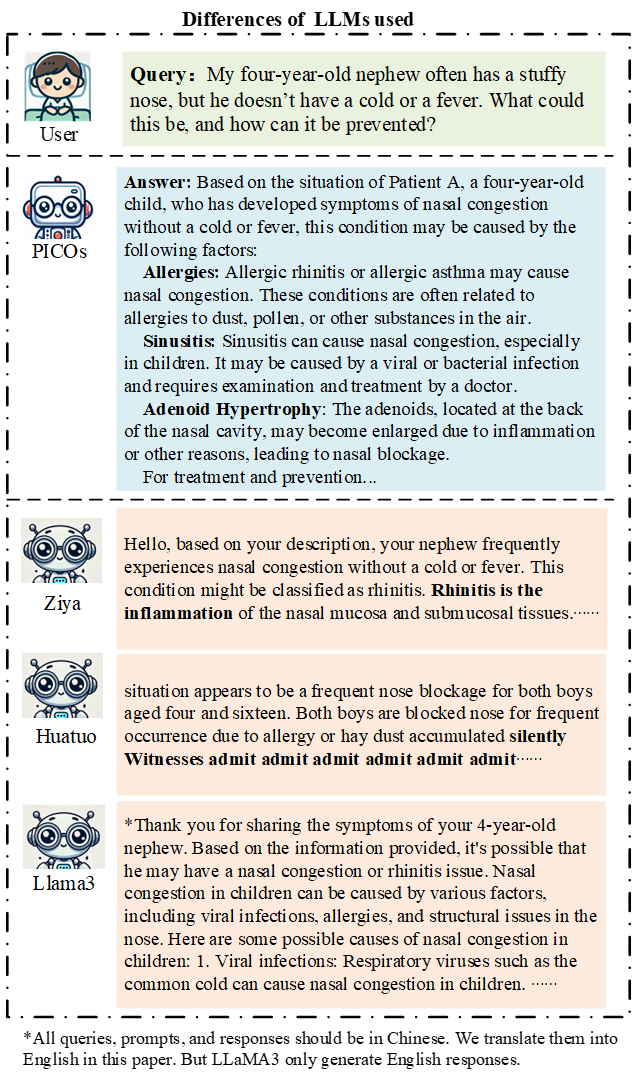}
    \caption{Examples of the pertinence of PICOs method generated. ZIYA tends to directly replicate the evidence. Huatuo2 tends to generate Meaningless repetition. LLaMA3 generates a good response but in the wrong language.}
    \label{fig:7}
\end{figure}

Secondly, we utilize an extractive model approach called LLaMA-index~\cite{Liu_LlamaIndex_2022} for RAG. We vectorize the medical guidelines datasets mentioned above as an index. The retrieval is performed by comparing the similarity between the search query and the evidence. The retrieved content is then input into the tested LLM as a reference for answering medical questions.

Next, for query rewriting and response, we both select the Baichuan2-13B model because it would be impractical and cumbersome to require medical institutions to deploy multiple models. We test various models, including LLaMA2~\cite{touvron2023llama}, LLaMA3~\cite{dubey2024llama}, BaiChuan2~\cite{yang2023baichuan}, Huatuo2~\cite{zhang2023huatuogpt} and Ziya-LLaMA-13B~\cite{fengshenbang}. Ultimately, we select the BaiChuan2-13B model for its best performance. As shown in Figure \ref{fig:7}, We find Huatuo and ZiYa models are hard to fully grasp our intention to expand the query and instead directly answer the questions frequently. At the response step, Huatuo2 tends to generate meaningless repetition until the end of the response, while ZIYA tends to directly replicate the content from the evidence. The LLaMA3 model lacks sufficient Chinese medical knowledge, resulting in a response in the wrong language. Consequently, we choose the BaiChuan2-13B model for the query expansion and normalization process. Also, we test some strong commercial LLMs like GPT-4o and GPT-4o-mini~\cite{openai2024gpt4}. As a generative model, GPT-4o is too expensive and tends to adopt a cautious approach when answering medical questions. Although this strategy often results in high scores during the evaluations, it lacks practical significance in providing meaningful responses.

Finally, in the process of selecting suitable evaluation models, We test the performance of various LLMs as evaluators and ultimately select BaiChuan2-13B and GPT-4o-mini as the models in Method A and Method B. As explained in Section III, The results obtained from Method A do not represent absolute accuracy but rather provide a measure of whether the response has covered the gold answer. During our experiments, we observe the BaiChuan2-13B as the evaluator strictly adheres to the prompt of comparing the gold answer. In contrast, GPT-4o-mini, with its more extensive knowledge base, focuses more on whether the response is substantively correct. Although we give the GPT-4o-mini the gold answer, the model is more inclined to make judgments based on its internal knowledge.  Therefore, as shown in Figure \ref{fig:6}, we believe that Method A (evaluated by BaiChuan2-13B) better reflects the relevance of the model's response. Method B (evaluated by GPT-4o-mini) provides a better indication of whether the content was comprehensive and accurate.

\subsection{Performance of PICOs-RAG}\label{4.2}
In this study, we use direct input from users as the control group and our PICOs-RAG reinforcement method as the experimental group to evaluate its effectiveness in identifying medical questions. We extract 500 queries randomly, covering a range of topics including medical treatment, diagnosis, and fee consultations. As shown in Table \ref{tab1}, after applying our PICOs method, the responses become clearer and more relevant to the query. However, the GPT-4o methods rise to such a high accuracy at 56.4\% evaluated by Method B, while Method A is 10.2\% lower than the PICOs. Also, It’s worth highlighting that Method B treats any output in a non-Chinese language as incorrect. As a result, the accuracy of the LLaMA model was notably lower when assessed using Method B.

\begin{table}[htbp]
\caption{behavior of PICOs and BASELINES}
\begin{center}
\begin{tabular}{c c c}

\hline
\hline
\textbf{LLMs to evaluate}&\textbf{Method A}&\textbf{Method B}\\
\hline
\textbf{Baichuan2-13B }  & 78.6\%& 30.4\%  \\
\textbf{Baichuan2-13B PICOs} & \textbf{84.8\% }& \textbf{39.2\%} \\
\hline

\textbf{Gpt-4o-mini}  & 66.6\%& 19.0\%  \\
\textbf{Gpt-4o-mini PICOs} & \textbf{74.6\%} & \textbf{56.4\%} \\
\hline
\textbf{Llama3-7B} & 42.8\% & \textbf{18.6\%} \\
\textbf{Llama3-7B PICOs} &\textbf{ 78.0\%} & 6.2\% \\

\hline
\hline
\end{tabular}
\label{tab1}
\end{center}
\end{table}

\begin{figure}
    \centering
    \includegraphics[width=1.0\linewidth]{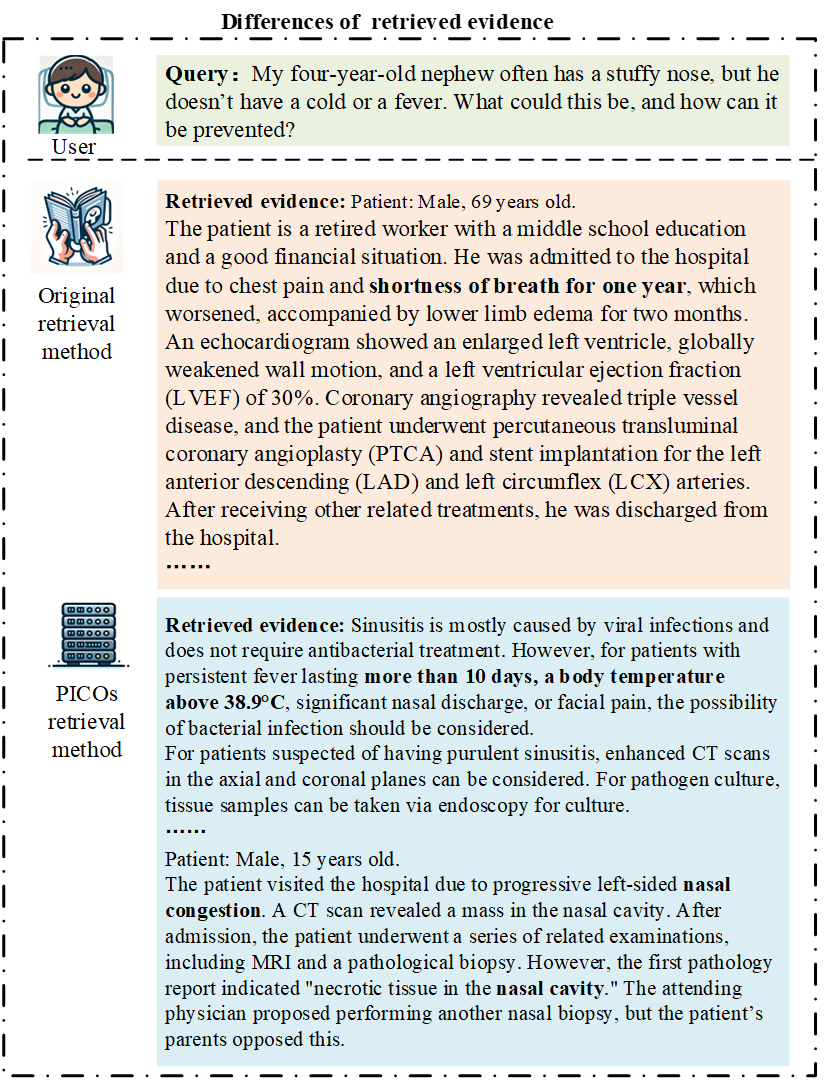}
    \caption{PICOs can retrieve better evidence than the original RAG method. Without the PICOs rewriting, the original rewriting retrieves too much noise and the evidence may be unhelpful for the generator}
    \label{fig:8}
\end{figure}

As shown in Figure \ref{fig:8}, we give out example of the evidence retrieved by PICOs-RAG and the current RAG method. The original method retrieves a 69-year-old male patient case as evidence. Moreover, the evidence retrieved in the case pertains to pneumonia, which is distantly related to the query about nasal congestion. In contrast, our PICOs-RAG method successfully extracts more relevant symptoms, such as sinusitis and allergies in young patients, which are more closely aligned with the query. The PICOs-RAG approach leads to more focused evidence retrieval and responses that more effectively address the key points.

Regarding the accuracy of each component, we obtain a dataset that includes the classification results of medical disciplines from the model itself. For medical texts, the classification of user input is often considered a priory knowledge. However, to address the scenario where fully open-ended questions might lack clear user classification, we train an input question classification model using the BioBERT model. Following the dataset setting, we classify the inputs into 22 medical disciplines and achieved an accuracy of 70.2\% on this dataset. This result demonstrates that the classification task can be successfully finished anyway.

\subsection{Ablation Experiment}\label{5.2}
\begin{table}[htbp]
\caption{Ablation experiments with baichuan2-13B.}
\begin{center}
\begin{tabular}{c c c}
\hline
\hline
\textbf{Methods to evaluate}&\textbf{Method A}&\textbf{Method B}\\
\hline
\textbf{PICOs} & 84.8\%& 39.4\%  \\
\textbf{w/o classification} & 72.8\%& 41.0\% \\
\textbf{w/o expansion} & 64.6\%& 22.6\% \\
\textbf{w/o PICO} & 70.4\%& 21.8\% \\
\hline
\textbf{w/ PIO} & 82.6\%&  39.2\% \\

\hline
\hline
\end{tabular}
\label{tab2}
\end{center}
\end{table}
As shown in Table \ref{tab2}, we conduct four different ablation experiments to validate the importance of each component of our method:
\begin{itemize}
    \item  \textbf{w/o classification:} First, we conduct a test on the classification of user inputs. We exclude the classification results from the prompt responsible for sentence expansion and normalization. The performance comparison is shown in Table \ref{tab2}. Under the evaluation Method B, we observe a slight improvement in the accuracy of response. However, the accuracy evaluated by Method A significantly decreases. We analyze the reason for this phenomenon and conclude that the lack of discipline-specific restrictions in the query leads the model to expand on the question in a more generalized manner. As a result, the responses generated by the LLM are more comprehensive but lack focus. This leads the evaluation model to believe that the response more thoroughly addresses a more comprehensive response. In contrast, Method A considers the response to lack the professional interpretation of discipline knowledge compared to the gold answer. Thereby the method without classification fails to accurately meet the user's intent. Consequently, such a huge decrease in Method A is Contrary to the main argument in this paper.
    \item    \textbf{w/o expanding:} We question whether expanded queries are easier to answer compared to original inputs. Therefore, we directly apply the PICOs extraction method to the original queries and then pose them to the generative model. This approach, without the LLM expanded queries, results in PICOs extraction that is not sufficiently targeted just as shown in Figure \ref{fig:8}. Consequently, the model's output remains a huge decrease in both evaluation methods shown in Table \ref{tab2}.

    \item   \textbf{w/o PICOs:} Next, we conduct ablation experiments on the performance of the PICO extraction method. As shown in the fourth line of Table \ref{tab2}, directly submitting the expanded information to the retrieval component decreased retrieval accuracy due to the noise introduced. This noise leads to the identification of many irrelevant indexes, which has a significantly negative impact on the final question-answering performance.
    \item   \textbf{w PIO:} Finally, we also test the type of PICO extracted components. Currently, there are various classification methods for the components of clinical articles. We conclude the primary distinction is whether to apply Comparison (C) and Study design (S) \cite{jin2018pico,methley2014pico}. Since our queries are sourced from online medical Q\&A platforms, the query does not require evidence grading. Therefore, this experiment focuses on validating which method, PICO or PIO, is more efficient. As shown in the last row of Table \ref{tab2}, the results of the two methods are relatively close. Therefore, we ultimately select the more effective PICO method.

\end{itemize}

\section{conclusion}
 In the field of EBM, there is an increasing need for robust automated tools to assist doctors in medical tasks. However, practical medical scenarios present numerous challenges and limitations for the LLMs application. To address this challenge, we propose a query rewriting and PICO extraction strategy aimed at improving the performance of RAG in medical natural question-answering environments. We conduct practical tests on our method and validate its effectiveness. We hope this work will assist researchers in the medical field, facilitating the safer and more effective deployment of LLMs in medical applications.
\section{Ethical Statement }
All datasets used in this study are open source, ensuring there are no ethical concerns regarding patient information disclosure.

\printbibliography

@article{zhang2023siren,
  title={Siren's Song in the AI Ocean: A Survey on Hallucination in Large Language Models},
  author={Zhang, Yue and Li, Yafu and Cui, Leyang and Cai, Deng and Liu, Lemao and Fu, Tingchen and Huang, Xinting and Zhao, Enbo and Zhang, Yu and Chen, Yulong and others},
  journal={arXiv preprint arXiv:2309.01219},
  year={2023}
}

@article{singhal2023towards,
  title={Towards expert-level medical question answering with large language models},
  author={Singhal, Karan and Tu, Tao and Gottweis, Juraj and Sayres, Rory and Wulczyn, Ellery and Hou, Le and Clark, Kevin and Pfohl, Stephen and Cole-Lewis, Heather and Neal, Darlene and others},
  journal={arXiv preprint arXiv:2305.09617},
  year={2023}
}

@article{huanglei2023survey,
  title={A survey on hallucination in large language models: Principles, taxonomy, challenges, and open questions},
  author={Huang, Lei and Yu, Weijiang and Ma, Weitao and Zhong, Weihong and Feng, Zhangyin and Wang, Haotian and Chen, Qianglong and Peng, Weihua and Feng, Xiaocheng and Qin, Bing and others},
  journal={arXiv preprint arXiv:2311.05232},
  year={2023}
}

@article{zhang2023huatuogpt,
  title={HuatuoGPT, towards Taming Language Model to Be a Doctor},
  author={Zhang, Hongbo and Chen, Junying and Jiang, Feng and Yu, Fei and Chen, Zhihong and Li, Jianquan and Chen, Guiming and Wu, Xiangbo and Zhang, Zhiyi and Xiao, Qingying and others},
  journal={arXiv preprint arXiv:2305.15075},
  year={2023}
}

@article{clusmann2023future,
  title={The future landscape of large language models in medicine},
  author={Clusmann, Jan and Kolbinger, Fiona R and Muti, Hannah Sophie and Carrero, Zunamys I and Eckardt, Jan-Niklas and Laleh, Narmin Ghaffari and L{\"o}ffler, Chiara Maria Lavinia and Schwarzkopf, Sophie-Caroline and Unger, Michaela and Veldhuizen, Gregory P and others},
  journal={Communications Medicine},
  volume={3},
  number={1},
  pages={141},
  year={2023},
  publisher={Nature Publishing Group UK London}
}

@article{hu2023towards,
  title={Towards precise PICO extraction from abstracts of randomized controlled trials using a section-specific learning approach},
  author={Hu, Yan and Keloth, Vipina K and Raja, Kalpana and Chen, Yong and Xu, Hua},
  journal={Bioinformatics},
  volume={39},
  number={9},
  pages={btad542},
  year={2023},
  publisher={Oxford University Press}
}

@article{friedman1999natural,
  title={Natural language processing and its future in medicine},
  author={Friedman, Carol and Hripcsak, George and others},
  journal={Acad Med},
  volume={74},
  number={8},
  pages={890--5},
  year={1999}
}

@article{friedman2013natural,
  title={Natural language processing: state of the art and prospects for significant progress, a workshop sponsored by the National Library of Medicine},
  author={Friedman, Carol and Rindflesch, Thomas C and Corn, Milton},
  journal={Journal of biomedical informatics},
  volume={46},
  number={5},
  pages={765--773},
  year={2013},
  publisher={Elsevier}
}

@article{nadkarni2011natural,
  title={Natural language processing: an introduction},
  author={Nadkarni, Prakash M and Ohno-Machado, Lucila and Chapman, Wendy W},
  journal={Journal of the American Medical Informatics Association},
  volume={18},
  number={5},
  pages={544--551},
  year={2011},
  publisher={BMJ Group BMA House, Tavistock Square, London, WC1H 9JR}
}

@article{subbiah2023next,
  title={The next generation of evidence-based medicine},
  author={Subbiah, Vivek},
  journal={Nature medicine},
  volume={29},
  number={1},
  pages={49--58},
  year={2023},
  publisher={Nature Publishing Group US New York}
}

@article{armeni2022digital,
  title={Digital twins in healthcare: is it the beginning of a new era of evidence-based medicine? A critical review},
  author={Armeni, Patrizio and Polat, Irem and De Rossi, Leonardo Maria and Diaferia, Lorenzo and Meregalli, Severino and Gatti, Anna},
  journal={Journal of Personalized Medicine},
  volume={12},
  number={8},
  pages={1255},
  year={2022},
  publisher={MDPI}
}

@article{vaid2024generative,
  title={Generative Large Language Models are autonomous practitioners of evidence-based medicine},
  author={Vaid, Akhil and Lampert, Joshua and Lee, Juhee and Sawant, Ashwin and Apakama, Donald and Sakhuja, Ankit and Soroush, Ali and Lee, Denise and Landi, Isotta and Bussola, Nicole and others},
  journal={arXiv preprint arXiv:2401.02851},
  year={2024}
}

@book{borenstein2021introduction,
  title={Introduction to meta-analysis},
  author={Borenstein, Michael and Hedges, Larry V and Higgins, Julian PT and Rothstein, Hannah R},
  year={2021},
  publisher={John Wiley \& Sons}
}

@article{ngusie2023effect,
  title={The effect of capacity building evidence-based medicine training on its implementation among healthcare professionals in Southwest Ethiopia: a controlled quasi-experimental outcome evaluation},
  author={Ngusie, Habtamu Setegn and Ahmed, Mohammadjud Hasen and Mengiste, Shegaw Anagaw and Kebede, Mihretu M and Shemsu, Shuayib and Kanfie, Shuma Gosha and Kassie, Sisay Yitayih and Kalayou, Mulugeta Hayelom and Gullslett, Monika Knudsen},
  journal={BMC medical informatics and decision making},
  volume={23},
  number={1},
  pages={172},
  year={2023},
  publisher={Springer}
}

@article{singhal2023large,
  title={Large language models encode clinical knowledge},
  author={Singhal, Karan and Azizi, Shekoofeh and Tu, Tao and Mahdavi, S Sara and Wei, Jason and Chung, Hyung Won and Scales, Nathan and Tanwani, Ajay and Cole-Lewis, Heather and Pfohl, Stephen and others},
  journal={Nature},
  volume={620},
  number={7972},
  pages={172--180},
  year={2023},
  publisher={Nature Publishing Group}
}

@article{mao2024rafe,
  title={RaFe: Ranking Feedback Improves Query Rewriting for RAG},
  author={Mao, Shengyu and Jiang, Yong and Chen, Boli and Li, Xiao and Wang, Peng and Wang, Xinyu and Xie, Pengjun and Huang, Fei and Chen, Huajun and Zhang, Ningyu},
  journal={arXiv preprint arXiv:2405.14431},
  year={2024}
}

@article{yang2023baichuan,
  title={Baichuan 2: Open large-scale language models},
  author={Yang, Aiyuan and Xiao, Bin and Wang, Bingning and Zhang, Borong and Bian, Ce and Yin, Chao and Lv, Chenxu and Pan, Da and Wang, Dian and Yan, Dong and others},
  journal={arXiv preprint arXiv:2309.10305},
  year={2023}
}

@article{dubey2024llama,
  title={The Llama 3 Herd of Models},
  author={Dubey, Abhimanyu and Jauhri, Abhinav and Pandey, Abhinav and Kadian, Abhishek and Al-Dahle, Ahmad and Letman, Aiesha and Mathur, Akhil and Schelten, Alan and Yang, Amy and Fan, Angela and others},
  journal={arXiv preprint arXiv:2407.21783},
  year={2024}
}

@article{touvron2023llama,
  title={Llama 2: Open foundation and fine-tuned chat models},
  author={Touvron, Hugo and Martin, Louis and Stone, Kevin and Albert, Peter and Almahairi, Amjad and Babaei, Yasmine and Bashlykov, Nikolay and Batra, Soumya and Bhargava, Prajjwal and Bhosale, Shruti and others},
  journal={arXiv preprint arXiv:2307.09288},
  year={2023}
}

@article{fengshenbang,
  author    = {Jiaxing Zhang and Ruyi Gan and Junjie Wang and Yuxiang Zhang and Lin Zhang and Ping Yang and Xinyu Gao and Ziwei Wu and Xiaoqun Dong and Junqing He and Jianheng Zhuo and Qi Yang and Yongfeng Huang and Xiayu Li and Yanghan Wu and Junyu Lu and Xinyu Zhu and Weifeng Chen and Ting Han and Kunhao Pan and Rui Wang and Hao Wang and Xiaojun Wu and Zhongshen Zeng and Chongpei Chen},
  title     = {Fengshenbang 1.0: Being the Foundation of Chinese Cognitive Intelligence},
  journal   = {CoRR},
  volume    = {abs/2209.02970},
  year      = {2022}
}

@article{alam2023automated,
  title={Automated clinical knowledge graph generation framework for evidence based medicine},
  author={Alam, Fakhare and Giglou, Hamed Babaei and Malik, Khalid Mahmood},
  journal={Expert Systems with Applications},
  volume={233},
  pages={120964},
  year={2023},
  publisher={Elsevier}
}

@article{ma2023query,
  title={Query rewriting for retrieval-augmented large language models},
  author={Ma, Xinbei and Gong, Yeyun and He, Pengcheng and Zhao, Hai and Duan, Nan},
  journal={arXiv preprint arXiv:2305.14283},
  year={2023}
}

@article{methley2014pico,
  title={PICO, PICOS and SPIDER: a comparison study of specificity and sensitivity in three search tools for qualitative systematic reviews},
  author={Methley, Abigail M and Campbell, Stephen and Chew-Graham, Carolyn and McNally, Rosalind and Cheraghi-Sohi, Sudeh},
  journal={BMC health services research},
  volume={14},
  number={1},
  pages={1--10},
  year={2014},
  publisher={Springer}
}

@article{he2019applying,
  title={Applying deep matching networks to Chinese medical question answering: a study and a dataset},
  author={He, Junqing and Fu, Mingming and Tu, Manshu},
  journal={BMC medical informatics and decision making},
  volume={19},
  pages={91--100},
  year={2019},
  publisher={Springer}
}

@software{Liu_LlamaIndex_2022,
author = {Liu, Jerry},
doi = {10.5281/zenodo.1234},
month = {11},
title = {{LlamaIndex}},
url = {https://github.com/jerryjliu/llama_index},
year = {2022}
}

@misc{openai2024gpt4,
  author = {OpenAI},
  title = {GPT-4},
  year = {2024},
  url = {https://openai.com/gpt-4},
  note = {Accessed: 2024-08-16}
}

@inproceedings{jin2018pico,
  title={Pico element detection in medical text via long short-term memory neural networks},
  author={Jin, Di and Szolovits, Peter},
  booktitle={Proceedings of the BioNLP 2018 workshop},
  pages={67--75},
  year={2018}
}
\end{document}